\documentclass[journal]{IEEEtran}
\usepackage{times}
\usepackage{mathrsfs}
\usepackage{epsfig}
\usepackage{graphicx}
\usepackage{amsmath}
\usepackage{amssymb}
\usepackage{graphicx}
\usepackage{cite}
\usepackage{enumerate}
\usepackage{cases}
\usepackage{multirow}
\usepackage{verbatim}
\usepackage{amssymb}
\usepackage{CJK}
\usepackage{algorithm}
\usepackage{algorithmicx}
\usepackage{algpseudocode}
\usepackage{color}
\usepackage{bm}
\usepackage{booktabs}
\usepackage{amssymb}

\usepackage[pagebackref=true,breaklinks=true,letterpaper=true,colorlinks,bookmarks=false]{hyperref}

\usepackage{color}
\usepackage{tikz}
\usepackage{ifthen}
\usepackage{bm}
\usepackage{multirow}
\def\triplets(#1,#2,#3){#1\quad#2\quad#3}

\usepackage{enumitem}

\usepackage{booktabs}

\newcommand{\etal}{\textit{et al}.}
\newcommand{\ie}{\textit{i}.\textit{e}.}
\newcommand{\eg}{\textit{e}.\textit{g}.}
\newcommand{\etc}{\textit{etc}}

\begin{document}

\title{Multi-Projection Fusion and Refinement Network for Salient Object Detection in 360$^{\circ }$ Omnidirectional Image}

\author
{
Runmin Cong,~\IEEEmembership{Member,~IEEE,} Ke Huang, Jianjun Lei,~\IEEEmembership{Senior Member,~IEEE,} \\Yao Zhao,~\IEEEmembership{Fellow,~IEEE,} Qingming Huang,~\IEEEmembership{Fellow,~IEEE,} and Sam Kwong,~\IEEEmembership{Fellow,~IEEE}

}

% paper header
\markboth{IEEE Transactions on Neural Networks and Learning Systems}
{Shell \MakeLowercase{\textit{et al.}}: Bare Demo of IEEEtran.cls for IEEE Journals}
\maketitle

\begin{abstract}
Salient object detection (SOD) aims to determine the most visually attractive objects in an image. With the development of virtual reality technology, 360$^{\circ }$ omnidirectional image has been widely used, but the SOD task in 360$^{\circ }$ omnidirectional image is seldom studied due to its severe distortions and complex scenes. In this paper, we propose a Multi-Projection Fusion and Refinement Network (MPFR-Net) to detect the salient objects in 360$^{\circ }$ omnidirectional image. Different from the existing methods, the equirectangular projection image and four corresponding cube-unfolding images are embedded into the network simultaneously as inputs, where the cube-unfolding images not only provide supplementary information for equirectangular projection image, but also ensure the object integrity of the cube-map projection. In order to make full use of these two projection modes, a Dynamic Weighting Fusion (DWF) module is designed to adaptively integrate the features of different projections in a complementary and dynamic manner from the perspective of inter and intra features. Furthermore, in order to fully explore the way of interaction between encoder and decoder features, a Filtration and Refinement (FR) module is designed to suppress the redundant information between the feature itself and the feature. Experimental results on two omnidirectional datasets demonstrate that the proposed approach outperforms the state-of-the-art methods both qualitatively and quantitatively. The code and results can be found from the link of \url{https://rmcong.github.io/proj\_MPFRNet.html}.
\end{abstract}


\IEEEpeerreviewmaketitle

\section{Introduction} \label{sec1}
\begin{figure}[t]
 	\centering
 	\includegraphics[width=\columnwidth]{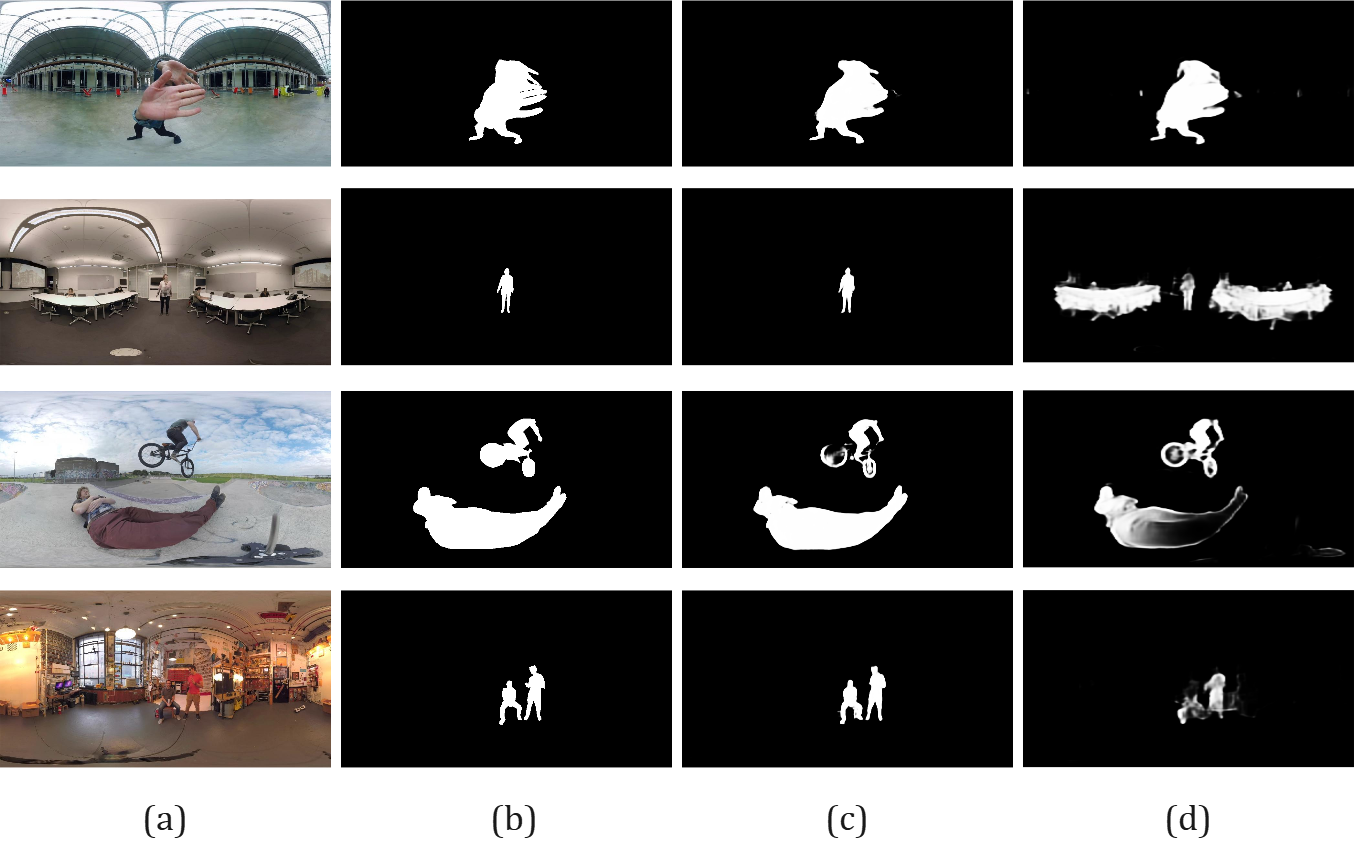}
 	\caption {Examples of the SOD in 360$^{\circ }$ omnidirectional images. (a) 360$^{\circ }$ omnidirectional images. (b) Ground truthes. (c) Results of our proposed method. (d) Result of ITSD \cite{zhou2020interactive}, which is a state-of-the-art SOD method for 2D images.
 	}
 	\label{fig1}
 \end{figure}

\IEEEPARstart{S}{ALIENT} object detection (SOD) aims to locate the most visually interesting and attractive objects/regions in an image \cite{crm2019tcsvt}, which is inspired by the human visual attention mechanism and has a wide range of applications in many fields, such as scene classification \cite{zhang2014saliency}, image captioning \cite{xu2015show}, quality assessment \cite{crm/SPIC21/underwaterIQA}, feature extraction \cite{tnnls3,crm/CVPR21/depthSR}, image retrieval \cite{gao2015database}, video segmentation \cite{wang2015saliency}\cite{wang2017saliency}, image cropping \cite{wang2018deep}\cite{wang2016stereoscopic}, object recognition \cite{ren2013region} and tracking \cite{hong2015online}. According to the different inputs, SOD models can be further divided into RGB SOD \cite{tnnls1,tnnls2,hou2017deeply,chen2020global,qin2019basnet,feng2019attentive,DBLP:conf/iccv/00380ZH19,wang2019salient,wang2019iterative,DBLP:conf/nips/ZhangTH20,qin2020u2,liu2019simple,crm/tcsvt22/weaklySOD}, RGB-D/RGB-T SOD \cite{tnnls4,lcy2020tc,crm2020tc,DPANet,crm/tip21/DynamicRGBDSOD,crm/acmmm21/CDINet,crm/tip22/CIRNet,crm/tmm22/TNet}, co-salient object detection (CoSOD) \cite{DBLP:journals/pami/ZhangMH17,DBLP:journals/tcsv/HanCLZ18,crm2019tmm,crm2018tip,crm2019tc,CoADNet,DBLP:conf/cvpr/FanLJZFC20,crm/tcyb22/glnet}, video SOD \cite{DBLP:journals/pami/WangSXCLB21,tnnls6,crm2019tip,wang2017video,wang2018salient,fan2019shifting,song2018pyramid,DBLP:journals/tip/LaiWSS20,DBLP:journals/tcsv/GuoWSSST20,DBLP:journals/tip/ChenWPFZQ21,crm/tetci22/PSNet}, remote sensing SOD \cite{crm2019tgrs,DAFNet,RRNet,crm/tcyb22/rsi}, \etc. Recently, with the development of hardware and VR applications, the input of the SOD task is no longer limited to traditional 2D images, and the SOD in 360$^{\circ }$ omnidirectional images has gradually emerged. 
The biggest difference between traditional 2D images and 360$^{\circ }$ omnidirectional images is the field of view (FOV). The 360$^{\circ }$ omnidirectional image can capture information from all directions, and its FOV covers the entire sphere, with a 360$^{\circ }$ FOV in both horizontal and vertical directions. Such images are becoming increasingly popular in the fields like virtual reality and robotics. Salient object detection in 360° omnidirectional image can be worked as a preprocessing step for many other tasks related to omnidirectional images, such as omnidirectional image coding \cite{luz2017saliency}, editing \cite{serrano2017movie}, stitching \cite{li2019attentive}, quality assessment \cite{zhou2021projection}, and navigation \cite{maugey2017saliency}. For example, since the resolution scale of 360$^{\circ }$ omnidirectional images is usually much larger than that of traditional 2D images, their transmission and storage costs are relatively high. With the help of salient object detection results, the effective bitrate allocation can be better achieved, thereby improving the efficiency of transmission and storage processes. In addition, with the popularity of 360$^{\circ }$ omnidirectional images in entertainment and photography, salient object detection in 360$^{\circ }$ omnidirectional images has also found more daily application scenarios, such as photo clustering for albums, VR live background replacement, \etc. However, as shown in Fig. \ref{fig1}, due to the severe distortion, shape deformation, large and complex scene of the 360$^{\circ }$ omnidirectional image, it is difficult to obtain satisfactory results by directly transplanting the 2D SOD model to the 360$^{\circ }$ omnidirectional image (\eg, Fig. \ref{fig1}(d)). Therefore, a dedicated network needs to be designed to solve the SOD task in 360$^{\circ }$ omnidirectional image.
Specifically, SOD in 360$^{\circ }$ omnidirectional image faces the following two main challenges.

First, the projection from sphere to plane will bring distortion, which is inevitable for any projection method. These distortions significantly increase the difficulty of salient object detection, easily resulting in missing detection, incomplete detection, wrong detection, and other problems. Equirectangular projection is the most commonly used sphere-to-plane projection and storage form for 360$^{\circ }$ omnidirectional image. This projection process is like forming a rectangular plane by keeping the distance between the latitudes of the Earth constant and stretching the longitudes into equidistant vertical parallel lines, which can effectively preserve the global information of an image. However, the extension of the poles from a point to a line will cause greater and greater distortion from the equator to the upper and lower boundaries. That way, the deep learning based SOD models for the traditional 2D image are difficult to identify the severely distorted salient object in the 360$^{\circ }$ omnidirectional image.
In addition to the equirectangular projection, cube-map projection \cite{maugey2017saliency} performs a perspective projection of the sphere on six surfaces of the cube, thereby obtaining six separate plane images. Compared to equirectangular projection, the distortion of cube-map projection is not so serious, but if the six planes are used alone, the object may be split into several parts. At present, there are some attempts to combine these two types of projection data to achieve SOD in 360$^{\circ }$ omnidirectional image. For example, Huang \etal \cite{huang2020fanet} proposed a Feature Adaption Network (FANet) for SOD in 360$^{\circ }$ omnidirectional image, which takes the equirectangular image and the six corresponding cube-map images as inputs simultaneously, aiming to leverage the respective advantages of these two types of features and achieve gratifying performance. In this way, the equirectangular projection provides global information, and the cube projection with less distortion complements it by providing structural guidance. However, it ignores the discontinuity caused by using the six cube surfaces separately, affecting the integrity and continuous expression of the object. 
Therefore, we consider unfolding the cube into a set of planar figures, where the cube surfaces are connected in a certain manner (\ie, four surfaces connected horizontally and three surfaces connected vertically) to maximize the objects continuity. Further, the equirectangular projection (EP) image and four cube-unfolding (CU) images are worked as the inputs of the SOD network, which not only fully explores the complementarity of the two projection modes, but also ensures the object integrity of the cube-map projection.

Another main challenge of the 360$^{\circ }$ SOD is that, since a 360$^{\circ }$ omnidirectional image covers an omnidirectional view, it usually leads to a large scene with much redundant information in the background, which interferes with the localization of salient objects.
With the redundant information of multi-source inputs, how to effectively realize the complementary fusion and filtering refinement of information is another key issue that must be faced and resolved in the implementation process. On the one hand, the two projection features have both information complementarity and content redundancy, and each plays a different role. Thus, indiscriminate fusion strategies (such as the concatenation-convolution, addition, \etc) will reduce the feature expression ability of the network. Because of this, we adaptively fuse the equirectangular and cube-unfolding encoder features through a Dynamic Weighting Fusion (DWF) module, in which different weights are assigned to intra and inter fusions. On the other hand, the positive effect of encoder features on feature decoding has been verified, but how to ensure the effectiveness and compactness of information transmission is still a problem worthy of in-depth discussion, especially for the multi-source inputs. To this end, we design a Filtration and Refinement (FR) module to suppress the redundant information between the feature itself and the feature, thereby achieving the feature selection and integration.

The main contributions of the proposed method are summarized as follows.
\begin{itemize}[noitemsep, topsep=0pt]
	\item A Multi-Projection Fusion and Refinement Network (MPFR-Net) is proposed to achieve salient object detection in 360$^{\circ }$ omnidirectional images, which introduces four cube-unfolding images through cube-map projection to supplement the corresponding equirectangular image.
	\item A Dynamic Weighting Fusion (DWF) module is designed to adaptively integrate the multi-projection features from the perspective of inter and intra features in a complementary and dynamic manner.
	\item A Filtration and Refinement (FR) module is presented to update the encoder and decoder features through the filtration scheme, and to suppress the redundant information in the fusion process through the refinement scheme.
\end{itemize}

%The rest of this paper is organized as follows. Section II reviews the related works of SOD models in conventional 2D images and 360$^{\circ }$ omnidirectional images. Section III introduces the proposed method in detail. The comprehensive experiments are presented in Section IV. Finally, the conclusion is drawn in Section V.
%-------------------------------------------------------------------------

\begin{figure*}[t]
	 \centering
	\includegraphics[width=\textwidth]{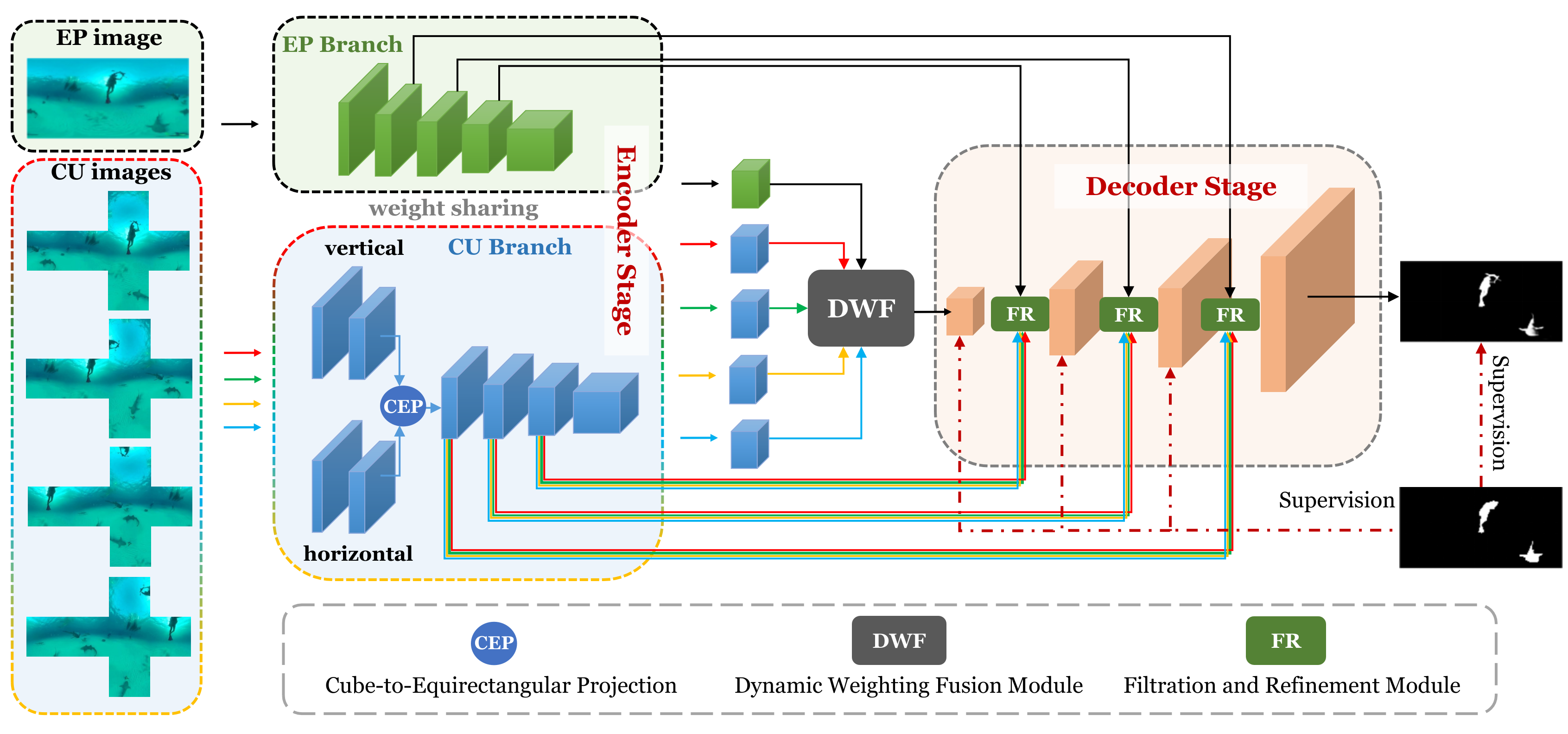}	
	\caption{The overall pipeline of the proposed \textit{MPFR-Net}. Equirectangular image and four cube-unfolding images are used as the inputs of the encoder-decoder network. The encoder is a weight-shared ResNet-50 composed of five convolutional blocks. In the feature decoding process, the equirectangular features and cube-unfolding features are firstly fused by the DWF module in a dynamic weighing manner. Then, in each level of decoding, the redundant information in features is filtered and refined by the FR module.
		}
	\label{fig2}
\end{figure*}
\section{Related Work}
%In this section, we discuss and summarize the related works of saliency detection techniques, including salient object detection models in conventional 2D images and saliency models in 360$^{\circ }$ omnidirectional images.
\subsection{Salient Object Detection in 2D Images}
Over the past decades, hundreds of salient object detection models on conventional 2D images have been proposed. %Initially, the hand-crafted features are widely used in the traditional methods \cite{borji2012exploiting,6619251,cheng2014global,liu2010learning,xie2012bayesian,cheng2013efficient}. 
Especially in recent years, the SOD method based on deep learning has won a very competitive performance \cite{tnnls1,tnnls2,hou2017deeply,chen2020global,qin2019basnet,feng2019attentive,wang2019salient,wang2019iterative,DBLP:conf/nips/ZhangTH20,qin2020u2,liu2019simple,zhao2015saliency,DBLP:conf/iccv/00380ZH19,wang2016saliency,deng2018r3net,zhang2018bi,zhang2018progressive,luo2017non}. 
%Hou \etal \cite{hou2017deeply} introduced short connections between shallower and deeper side-output layers, which takes full advantage of multi-level and multi-scale features.
%Chen \etal \cite{chen2020global} proposed a progressive aggregation network, which considers the global context information and fully integrated different level features.
%Qin \etal \cite{qin2019basnet} proposed a boundary-aware model to optimize the results of saliency detection.
Feng \etal \cite{feng2019attentive} also proposed a boundary-aware detection model with attentive feedback modules and a boundary-enhanced loss. 
In order to emphasize the importance of salient edges, Wang \etal \cite{wang2019salient} presented an essential pyramid attention structure and equipped an edge detection module to strengthen the boundary information. 
Wang \etal \cite{wang2019iterative} presented a detection method, which integrated both top-down and bottom-up saliency inference in an iterative manner. 
Qin\cite{qin2020u2} constructed a two-level nested U-structure network to capture multi contextual information, increasing the depth of the network without large computational cost.
%Zhao \etal \cite{zhao2015saliency} proposed a multi-context deep CNN model which integrates global context and local context to predict the saliency map. Wang \etal \cite{wang2016saliency} presented a recurrent fully convolutional network to refine saliency maps in steps.
In addition, several methods have been proposed to integrate features of multiple layers to utilize context information of different levels. 
%Deng \etal \cite{deng2018r3net} proposed a recurrent network equipped with residual refinement blocks to obtain more accurate saliency map. 
%Zhang \etal \cite{zhang2018bi} proposed a novel bi-directional message passing model to integrate multi-level features better. 
%In \cite{zhang2018progressive}, Zhang \etal~presented a novel progressive attention network that selectively integrates contextual information from multi-level features.
Chen \etal \cite{luo2017non} presented an accurate yet compact deep network, which proposed reverse attention to guide such side-output residual learning in a top-down manner.
However, these methods only perform well on conventional 2D images, which cannot be directly applied to omnidirectional images.

\subsection{Salient Object Detection in 360$^{\circ }$ Omnidirectional Images}
%In the saliency detection field, there are two different task branches, fixation prediction and salient object detection. The former simulates human visual attention in a coarse level, while the latter pixel-wisely segments the objects that grasp most of the attention in human vision. Due to the distortion of omnidirectional images, conventional saliency detection models applying to omnidirectional images cannot have well performance.

In recent years, with the rapid development of virtual reality technology and application, the studies on SOD in 360$^{\circ }$ omnidirectional images has gradually attracted people's attention.
%there have been some study of saliency detection in omnidirectional images, which mainly in the fixation prediction field.
Monroy \etal \cite{monroy2018salnet360} mapped an omnidirectional image into six patches and obtained saliency maps respectively through a CNN network, which are combined together by post-processing technique to deal with the heavy distortion. Assens \etal \cite{assens2017saltinet} introduced the temporal-aware representation of saliency information into the SOD model. For the 360$^{\circ }$ video, Zhang \etal \cite{zhang2018saliency} presented a novel spherical convolutional neural network, where the kernels were shared across all image patches on the sphere.
%However, few salient object detection models have been proposed compared to the fixation prediction.
Li \etal \cite{li2019distortion} proposed a distortion-adaptive module, which cuts the images into different patches and passes them through different convolutional kernels. In addition, a multi-scale contextual integration block is proposed to perceive and distinguish the objects in omnidirectional scenes. Huang \etal \cite{huang2020fanet} took equirectangular image and the corresponding six cube-map surfaces as inputs to extract multi-level features and fuse them together, which can ameliorate the performance of SOD in omnidirectional images.

%%By contrast, we propose a end-to-end framework to perceive the potentiality of the depth map, and incorporate the confidence score into the gated multi-modality attention module to simultaneously integrate cross-modal information and prevent the contamination from unreliable depth information.
% MAIN PLOT

%-------------------------------------------------------------------------

\section{Proposed Method}
In this section, we present the details of our designed SOD network on 360$^{\circ }$ omnidirectional image, which follows an encoder-decoder architecture. The pipeline is shown in Fig. \ref{fig2}. The equirectangular projection (EP) image and four cube-unfolding (CU) images through the cube-map projection are used together as input. The encoder is a weight-shared ResNet-50, which is used for feature extraction of EP and CU images. Note that, after the first two layers, a Cube-to-Equirectangular Projection (CEP) module is utilized to convert the cube-unfolding features into the equirectangular features, and then continue to extract the higher-level features. The learned multi-level and multi-source encoder features are embedded into the decoder stage to progressively restore the spatial resolution and generate the saliency prediction map. Specifically, a Dynamic Weighting Fusion (DWF) module is designed to adaptively fuse the equirectangular and the cube-unfolding encoder features from the perspectives of intra and inter features, thereby selecting the more effective and comprehensive feature representations. The Filtration and Refinement (FR) module is inserted a total of $3$ times between the two decoding blocks, which aims to filter the redundant information from the low-level features of different projections and enhance the high-level semantic features. The final predicted saliency map is generated under the joint supervision of the dominant loss function and all the side-output loss functions.
\begin{figure}[!t]
	 \centering
	\includegraphics[width=0.5\textwidth]{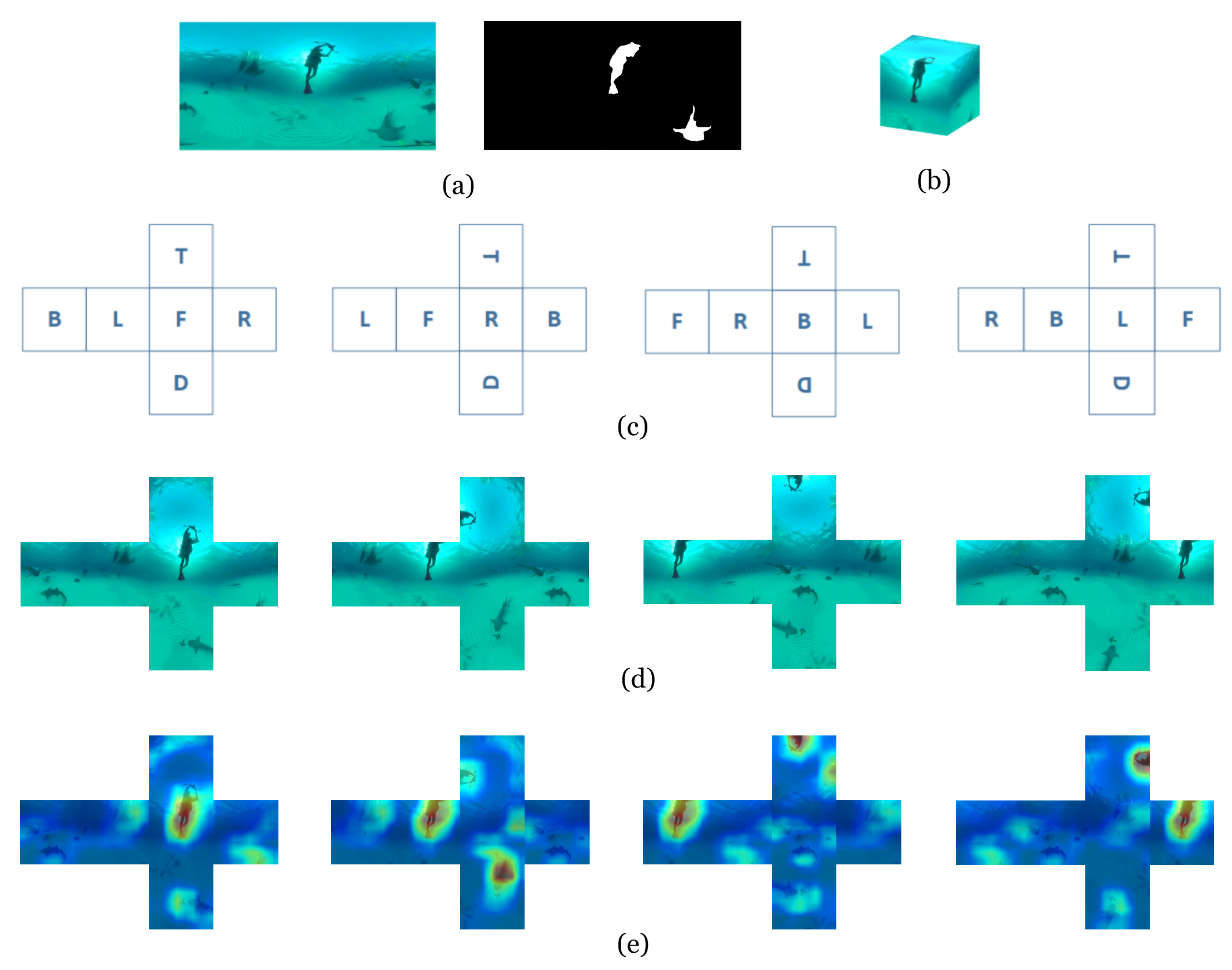}	
	
	\caption{Visualization of the cube unfolding. (a) Equirectangular projection image and the corresponding ground truth.  (b) Cube projection image. (c) Sketch maps of the cube unfolding in a 4-3 type. (d) The actual CU images centered on F, R, B, L respectively. (e) Visualization of the four cube unfolding features.
		}
	\label{fig3}
\end{figure}

In our network, the EP image and the corresponding four CU images are taken as inputs, and their correlations are reflected in the design of our DWF and FR modules. In the DWF module, the correlation of the EP and CU features is achieved through the inter and intra fusion of our design. Considering that the EP features can better perceive the global information and the CU features can describe the object attribute with less distortion, we achieve the inter fusion between each CU features and EP features, thereby automatically picking out the valuable parts of each of the two different projection features and generating the CU-induced fusion features. The intra fusion further considers the correlation and importance of different unfolding ways, and then learns a dynamic weight for each cube-unfolding manner to fuse the four CU-induced fusion features. Besides, in the FR module, we combine the EP and four CU encoder features with the decoder features, considering not only the associations among the EP and CU encoder features, but also their associations with the decoder features. Since the encoder features obtained by different projection methods have their own redundancy, we design a filtration scheme to filter out their redundancy and update the encoder features in a modulation manner. Then, the filtrated EP features and the four filtrated CU features are combined to strengthen the correlation among them and reduce the redundant information irrelevant to salient objects. Further combination with the decoder features is then done through a refinement scheme. In a word, we exploit complementarity of EP and CU features through the adaptive fusion in the DWF module, and use the EP encoder features and four CU encoder features together to generate encoding features in the FR module that complement the decoder features, thus further considering their correlation. 

\subsection{Cube Unfolding and Projection}
The EP image is one of the commonly used representations of the 360$^{\circ }$ omnidirectional image, which is also the original data provided by the existing SOD dataset. However, this form of representation has relatively large distortions, which deforms the shape of the salient object and affects the identification. Therefore, in addition to the EP image, the corresponding six surfaces of the cube-map projection are also used as one of the inputs of the network in \cite{huang2020fanet}, where EP image provides the global information, and the cube-map projection provides structural guidance as a supplement due to its less distortion. The experiments in this paper demonstrate that the combination of the two projection inputs can achieve better results than just using the EP image as input. However, inputting the six surfaces of the cube-map projection for feature extraction will damage the integrity and continuity of the objects, which is obviously not conducive to the salient object detection. To this end, we consider unfolding the cube into a planar figure in a certain arrangement to increase the possibility of making the salient object complete. 

%We choose the 4-3 type unfolding form, where the four surfaces are connected horizontally and the three surfaces are connected vertically.
There are a total of 11 manners to unfold a cube, and if there are two manners that can be overlapped after an arbitrary flip, we regard them as one. Among them, there are six types with four cube surfaces connected horizontally and the other two surfaces attached on the upper and lower side of the horizontal row; three types with three surfaces connected horizontally, one surface connected on one side of the horizontal row and two surfaces on the other; one type with two surfaces connected in the middle and two surfaces on each side; and the last type with only two rows of three surfaces each. Among the 11 kinds of cube unfolding methods, two are horizontally connected on four sides and vertical on three sides, which is called 4-3 type. This 4-3 type of unfolding manner ensures the longest possible continuity both horizontally and vertically, maximizing object integrity. By performing a 4-3 type unfolding centered on four surfaces (front, back, left, right), we can obtain four cube-unfolding (CU) images as inputs. In other words, in the vertical direction, the different unfolding methods centered on different surfaces build up different vertical connections, extending the vertical continuity in different directions. Since the real scene of a 360$^{\circ }$ omnidirectional image can only be fully displayed on a sphere, once converted from a sphere to a plane, the boundaries that more or less destroy the integrity of the object must be introduced. Moreover, the 360$^{\circ }$ omnidirectional images usually have vertically long-range objects, such as standing people, buildings, \etc. After converting a 360$^{\circ }$ omnidirectional image into a cube projection, these objects are often separated by the boundaries between the cube surfaces, thus destroying the integrity of the objects, which is very detrimental for detection. The 4-3 type of cube unfolding method respectively realizes the vertical connection in four directions, which significantly increases the integrity of the long-range object. Likewise, in the horizontal direction, the four surfaces of the cube are connected, and each surface undergoes a positional rotation from the center to the edge in the different unfolding methods. That is to say, if one makes two surfaces separated by left and right boundaries, they will be connected in all the other unfolding methods. Therefore, the continuity in the horizontal direction is also enhanced.

In the existing dataset for SOD in 360$^{\circ }$ omnidirectional images, the images are stored in the form of equirectangular projection. To convert such an image into a CU image requires two simple data preprocessing steps. In the first step, the EP image is resized to the size of ${H \times W}$, where $H$ is set to 1024 and $W$ is set to 512 in our experiments. Then, we transform an EP image (Fig. \ref{fig3}(a)) to cube-map projection image (Fig. \ref{fig3}(b)), producing six cube surfaces with the size of ${H/2 \times W/4}$. Let \{F, B, L, R, T, and D\} represent the front, behind, left, right, top, and down surfaces of a cube, where T and D correspond to the positions near the upper and lower boundaries in an EP image. In the second step, we unfold the cube with F, R, B, L as the centers respectively, and obtain four CU images in a 4-3 type as shown in Fig. \ref{fig3}(c-d), which are dented as ${{C}_{i}}$ $(i=\{1,2,3,4\})$.
As visible, the upper body and lower body of the diver are located on two different surfaces of the cube. If each cube surface is processed separately, the salient object (\ie, the diver) will be cut into two parts and lose continuity. Instead, when the cube is unfolded centered on F, this vertical diver can be preserved intact with less distortion. In this way, a longitudinal object has the opportunity to enter the neural network in a continuous manner regardless of whether it is in the middle or close to the boundary, while a transverse object has the opportunity to not be split by the left and right boundaries no matter where it is. 

In our network, the four CU images ${{C}_{i}}$ $(i=\{1,2,3,4\})$ together with the EP image $E$ are fed into the feature encoder sharing parameters to learn the multi-level features. In practice, each CU image is further split into a horizontal CU image of size $a \times 4a$ and a vertical CU image of size $3a \times a$, where $a$ means the side-length of the cube. This is because, if we take a CU image of the 4-3 type directly as input, it will result in large-scale zero-padding, which not only wastes computing resources, but also is not conducive to the localization of salient objects. Therefore, we split CU images into horizontal and vertical sub-images for feature extraction at the lower level, respectively. After the low-level extraction, the Cube-to-Equirectangular Projection (CEP) module is introduced to project the CU features into an equirectangular projection form, and then continue to extract the high-level semantic features. The purpose of this is to compensate for the slight discontinuity that occurs after the cube surfaces are stitched together, where the discontinuity is in detail rather than semantic. This is also conducive to the integration of the horizontal and vertical images. Besides, we provide the visualization features of the final encoder layer for four CU images in Fig. \ref{fig3}(e), which shows that the features corresponding to different cube unfolding methods have different degrees of emphasis on objects in the image. This opens up the possibility for complementary combinations of the four cube unfolding methods in the following modules.

\subsection{Dynamic Weighting Fusion Module}
\begin{figure}[!t]
 	\centering
 	\includegraphics[width=\columnwidth]{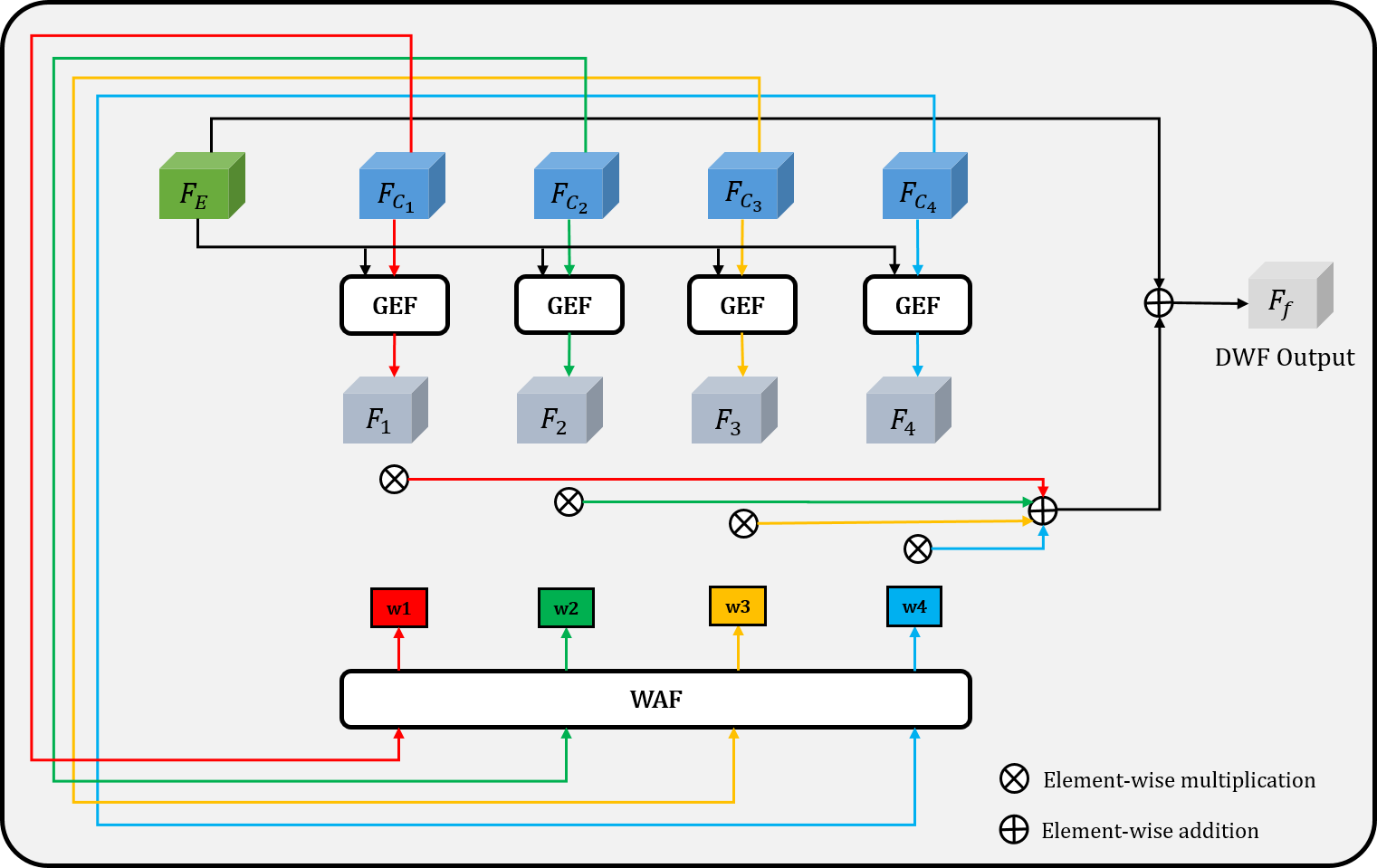}
 	\caption {The architecture of DWF module. GEF is the Gated Inter Fusion unit and WAF denotes the Weighted Intra Fusion unit.}
 	\label{fig4}
 \end{figure}

In fact, the equirectangular projection and cube projection are two ways to present the 360$^{\circ }$ omnidirectional image. As mentioned earlier, they have their own advantages and are complementary to a certain extent. On one hand, the EP features can better perceive the global information, but they also contain inevitable distortions in some objects. On the other hand, the CU features can describe the object attribute with less distortion and prevent the object truncation phenomenon to the greatest extent, thereby ensuring the integrity and continuity of the object perception. Therefore, by fusing the two types of features before feature decoding, the more comprehensive and effective features can be obtained, thereby achieving better decoding prediction. However, how to integrate them effectively and automatically is a problem worth exploring. Looking back at the features that need to be integrated, they can be divided into two categories: (1) the inter fusion between each CU features and EP features, thereby generating the CU-induced fusion features; (2) the intra fusion of the four CU-induced fusion features. The nomenclature of the inter and intra fusions depends on whether the features being fused are from the same projection type. The inter fusion aims to automatically pick out the valuable parts of each of the two different projection features, thereby generating the CU-induced fusion features. Differently, the intra fusion takes into account the different importance of different unfolding ways, which aims to fuse four CU-induced fusion features obtained in the inter fusion process. The main consideration for this is that different unfolding ways of cube-map projection images have a certain connection, and the distribution positions of the objects in each cube-map projection image are also different, so their importance to the salient object detection task is also different. Thus, we design the intra fusion to learn a dynamic weight for each cube-unfolding manner. In summary, the intra and inter fusion stages are integrated into the Dynamic Weighting Fusion (DWF) module to adaptively fuse the two projection modes in a complementary and dynamic way, including a Gated Inter Fusion (GEF) unit and a Weighted Intra Fusion (WAF) unit, as shown in Fig. \ref{fig4}.

\textbf{Gated Inter Fusion (GEF) unit}. It is known that EP features contain more global information to preserve the integrity of the scene, but also suffer from severe distortions. As for the CU features, they have less distortion with more accurate detail information to provide structural guidance as a supplement. Therefore, we propose the GEF unit to integrate each CU encoder features in $\{{{F}_{{{C}_{i}}}}\in \mathbb{R}^{C \times H \times W}\}_{i=1}^{4} $ 
and EP encoder features ${F}_{E}\in \mathbb{R}^{C \times H \times W}$ dynamically, picking out the valuable parts of each of the two different projection features.
%First, each CU encoder features in $\{{{F}_{{{C}_{i}}}}\in \mathbb{R}^{C \times H \times W}\}_{i=1}^{4} $ are adaptively fused with EP encoder features ${F}_{E}\in \mathbb{R}^{C \times H \times W}$ to achieve inter fusion through the GEF unit.
Concretely, we learn an importance map to evaluate how much CU information should be provided for the EP features. We first concatenate ${{F}_{E}}$ with each ${{F}_{{{C}_{i}}}}$ along the channel dimension, and then apply a SE block \cite{hu2018squeeze} for channel attention followed by a bottleneck convolutional layer and a sigmoid activation function, which can be formulated as:
\begin{equation}
{{P}_{i}}=\sigma(conv(SE([{{F}_{E}},{F}_{{{C}_{i}}}]))),
\end{equation}
where ${P}_{i}$ represents the contribution of the CU features ${F}_{{{C}_{i}}}$, and ${{F}_{E}}$ are the EP features and CU features, [·,·] is the channel concatenation, $\sigma$ denotes the sigmoid activation function and SE is a Squeeze-and-Excitation block. Thus, with the importance map, the CU-induced fusion features can be calculated by:
\begin{equation}
{{F}_{i}}={{P}_{i}}\circ {{F}_{C_i}}+(1-{{P}_{i}})\circ {{F}_{E}},
\end{equation}
where $\circ$ denotes the Hadamard multiplication, ${{F}_{i}}$ is the CU-induced fusion features, which is the final output of the GEF module.
% broadcasted along feature planes.

\textbf{Weighted Intra Fusion (WAF) unit}. 
The different roles of the four CU-induced fusion features determine that their contributions and importance to the final fusion features are also different.
Based on this, we design the WAF unit to learn the adaptive weights for the four CU-induced fusion features during the intra fusion, so that it can determine which unfolding method plays more important role for the equirectangular features, thereby highlighting the effective information and suppressing irrelevant redundancy. 
The original CU features $\{{{F}_{{{C}_{i}}}}\in \mathbb{R}^{C \times H \times W}\}_{i=1}^{4}$ are concatenated together and passed through a SE block \cite{hu2018squeeze} to obtain a vector $\alpha \in {\mathbb{R}^{4C\times 1\times 1}}$. Then, the vector is further cut into four vectors $\{{{\alpha }_{i}} \in \mathbb{R}^{C\times 1\times 1}\}_{i=1}^{4}$. After the normalization, we can obtain the dynamic weights ${{w}_{i}}$. The above process can be expressed as follows:
\begin{equation}
\alpha_{i}=Split(SE([F_{C_1},F_{C_2},F_{C_3},F_{C_4}])),
\end{equation}
\begin{equation}
{{w}_{i}}=\frac{Sum({{\alpha }_{i}})}{\sum\limits_{i=1}^{4}{Sum({{\alpha }_{i}})}},
\end{equation}
where $SE$ is the Squeeze-and-Excitation block \cite{hu2018squeeze}, $Split$ represents the operation of splitting a vector into four vectors, $[\cdot]$ represents the channel concatenation, and $Sum(\cdot)$ denotes the sum of all elements of a vector. Thus, we integrate four CU-induced features $\{{{F}_{i}}\}_{i=1}^{4}$ obtained by the GEF unit in an adaptive weighting manner, and combine them with the EP features ${{F}_{E}}$ to obtain the final fusion features $F_f$:
\begin{equation}
{{F}_{f}}=F_E \oplus \sum\limits_{i=1}^{4}{{{w}_{i}}\otimes{{F}_{i}}},
\end{equation}
where ${{w}_{i}}$ is the corresponding weight obtained by the WAF unit, $\oplus$ denotes the element-wise addition, and $\otimes$ is the element-wise multiplication broadcasted along feature planes.
%Finally, we add the EP features ${{F}_{E}}$ and the fused features ${{F}_{f}}$ to obtain the final fusion features $F_d$.

\subsection{Filtration and Refinement Module}
\begin{figure}[!t]
 	\centering
 	\includegraphics[width=\columnwidth]{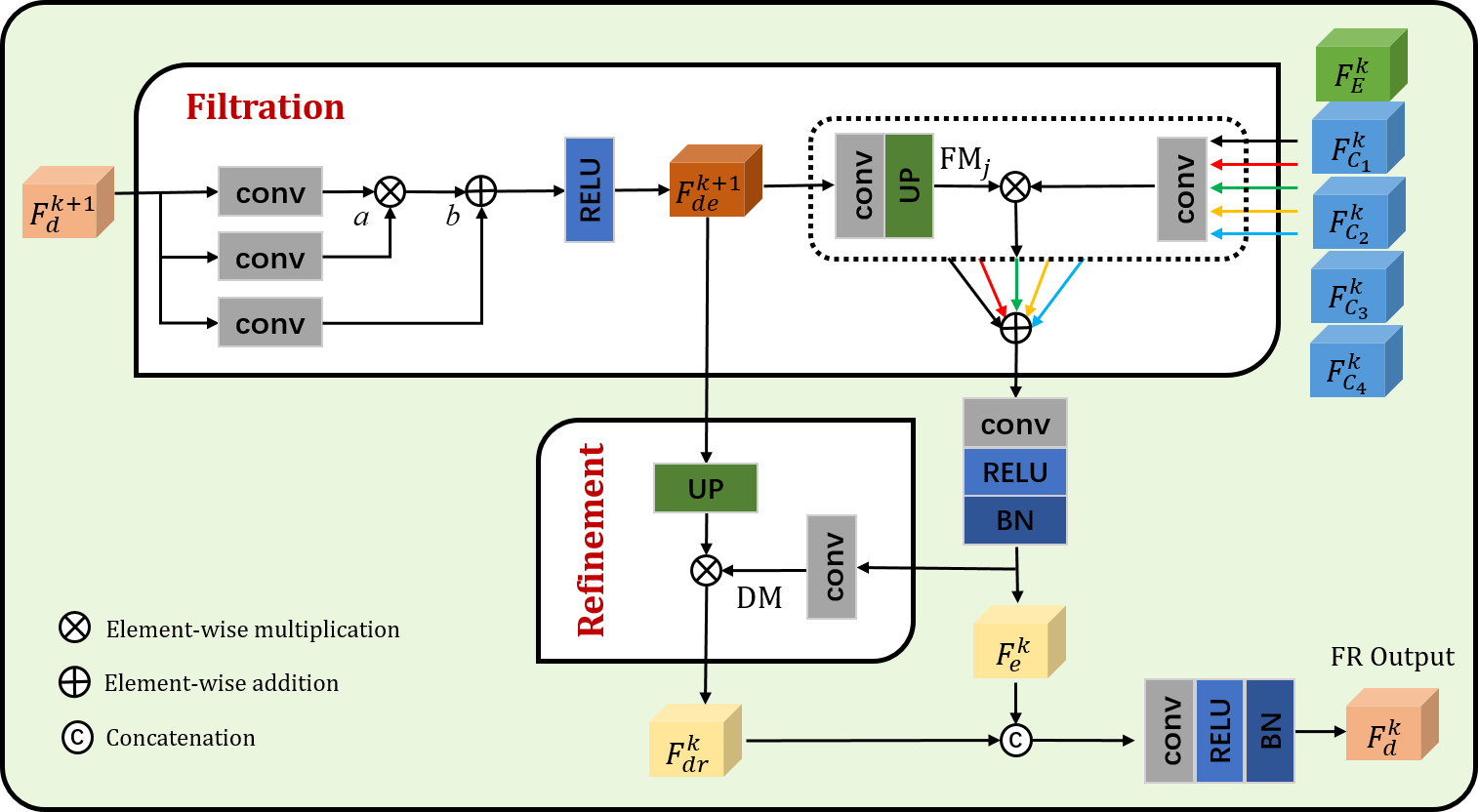}
 	\caption {The architecture of FR module.}
 	\label{fig5}
 \end{figure}

As is well-known, the combination of encoder and decoder features can provide more comprehensive saliency information, but these features also have a lot of redundant information, which will undoubtedly affect the expression ability of features. Depending on the source of redundant information, it can be manifested in two types. One is the redundancy of feature itself, which mainly refers to the redundant information on the spatial and channel of the encoder and decoder features. The other is the redundancy between features, referring to the redundancy between the encoder and decoder features in the process of feature fusion. Therefore, we design a Filtration and Refinement (FR) module to achieve feature selection and integration with a step-by-step strategy, which helps to suppress redundant information while remaining detail information relevant to the salient object, making the end-to-end training of the network more efficient. The detailed architecture of FR module is shown in Fig. \ref{fig5}. Taking the $k$-th FR module as an example ($k=\{2,3,4\}$), there are two types of six inputs in the FR module: the decoder features from the previous decoder layer $F_d^{k+1}$ and the encoder features from the corresponding encoder layers with different five inputs (\ie, $F_{E}^{k}$ and $\{F_{{{C}_{i}}}^{k}\}_{i=1}^{4}$). Note that, when $k=4$, the input of the previous decoder is the output of the DWF module.

To suppress the first type of redundancy, a filtration scheme is designed to update the encoder and decoder features. On one hand, inspired by \cite{chen2020global}, the decoder features are filtrated in a modulation manner. On the other hand, the encoder features focus more on the representation of general features without explicit saliency attributes, and thus contain a lot of redundancy. Therefore, to better adapt the encoder features to SOD task, we calculate a filtration mask $FM_j$ for each encoder features based on the filtrated decoder features, which encodes the saliency attribute. Then, the encoder features are multiplied by the corresponding filtration mask $FM_j$ to obtain the filtrated encoder features, thereby suppressing the redundancy of the encoder features.
To enhance the previous decoder features, we apply a $3\times 3$ convolution to reduce its channel number to $256$, and then utilize two convolutional layers to obtain the modulation parameters $a$ and $b$ for multiplication and addition, respectively. Thus, the enhanced decoder features can be derived by:
\begin{equation}
{{F}_{de}^{k+1}}=\delta (a\otimes F_{d}^{k+1}+b),
\end{equation}
where $\delta$ is the RELU function. As the encoder features focus more on the representation of general features without explicit saliency attributes, and thus contain a lot of redundancy. Therefore, to better adapt the five encoder features $F_{j}^{k}$, $j =\{{C}_{1},{C}_{2},{C}_{3},{C}_{4},E\}$  to SOD task, we apply a $3\times 3$ convolutional layer to the enhanced decoder features to obtain the corresponding filtration mask $\textrm{FM}_{j}$, which encodes the saliency attribute. Then, the encoder features are multiplied by the corresponding filtration mask $\textrm{FM}_{j}$ to obtain the filtrated encoder features $F_{e_j}^{f}$ , thereby suppressing the redundancy of the encoder features. The above process is represented by:
%\begin{equation}
%M_{j}^{f}=conv_{3\times3}(F_{d}^{f}),
%\end{equation}
\begin{equation}
F_{e_j}^{k}=\verb"UP"(\textrm{FM}_{j})\otimes conv_{3\times3}(F_{j}^{k}),
\end{equation}
where $\verb"UP"  $ denotes the upsampling operation. Finally, we can obtain the total filtrated encoder features $F_{e}^{k}$ through the addition operation:
%\begin{equation}
%F_{e}^{f}={{\delta }_{1}}({{\delta }_{2}}(conv(F_{e1}^{f}\oplus F_{e2}^{f}\oplus F_{e3}^{f}\oplus F_{e4}^{f}\oplus F_{e5}^{f}))),
%\end{equation}
\begin{equation}
F_{e}^{k}={{\delta }}(conv_{3\times3}(\sum_j F_{e_j}^{k})),
\end{equation}
where ${{\delta }}$ denotes the RELU activation function.

In addition, direct fusion of encoder and decoder features through the concatenation or addition operation will introduce a lot of unnecessary redundant information, corresponding to the second type of redundancy. Therefore, in the fusion process, it is essential to fully consider the role of the encoder and decoder features to suppress the achieve effective feature refinement. To this end, we design a refinement mechanism to suppress the redundancy of encoder and decoder features in the fusion process.
In order to be able to further highlight the important detail locations in the decoder features, we learn a detail mask $\textrm{DM}$ by performing convolution operations on the filtrated encoder features $F_{e}^{k}$, and use it to refine the filtrated decoder features:
%\begin{equation}
%\textrm{DM}=conv(F_{e}^{k}),
%\end{equation}
\begin{equation}
F_{dr}^{k}=\textrm{DM}\otimes \verb"UP"(F_{de}^{k+1}).
\end{equation}

After that, we can fuse the filtrated encoder features and the refined decoder features by:
\begin{equation}
F_{d}^{k}={{\delta }}(conv_{3\times3}([F_{e}^{k},F_{dr}^{k}]),
\end{equation}
where $F_{d}^{k}$ is the decoder features of the $k$-th layer embed into the next FR module.

\subsection{Supervision}
We use the binary cross-entropy as the loss function, which is calculated as:
\begin{equation}
L=-\frac{1}{W\times H}\sum\limits_{x=1}^{W}{\sum\limits_{y=1}^{H}{[{{G}_{xy}}\ln {{P}_{xy}}+(1-{{G}_{xy}})\ln (1-{{P}_{xy}})]}},
\end{equation}
where $W$ and $H$ denote the width and height of the EP image, respectively, ${{G}_{xy}}$ is the ground truth at location $(x,y)$, and ${{P}_{xy}}$ denotes the saliency prediction score at location $(x,y)$.

In the implementation, we set up supervision information for each decoder output, including one domination loss of the final saliency prediction and three side-output losses. The total loss function ${{L}_{total}}$ is defined as:
\begin{equation}
{{L}_{total}}={{L}_{dom}}+\sum\limits_{i=1}^{3}{{{\alpha }_{i}}L_{side}^{i}},
\end{equation}
where ${{L}_{dom}}$ denotes the domination loss of the final output, $L_{side}^{i}$ represents the side-output loss, and ${{\alpha }_{i}}$ is the weight of different side-output losses .

\begin{figure*}[t]
	\centering
	\includegraphics[width=1\textwidth]{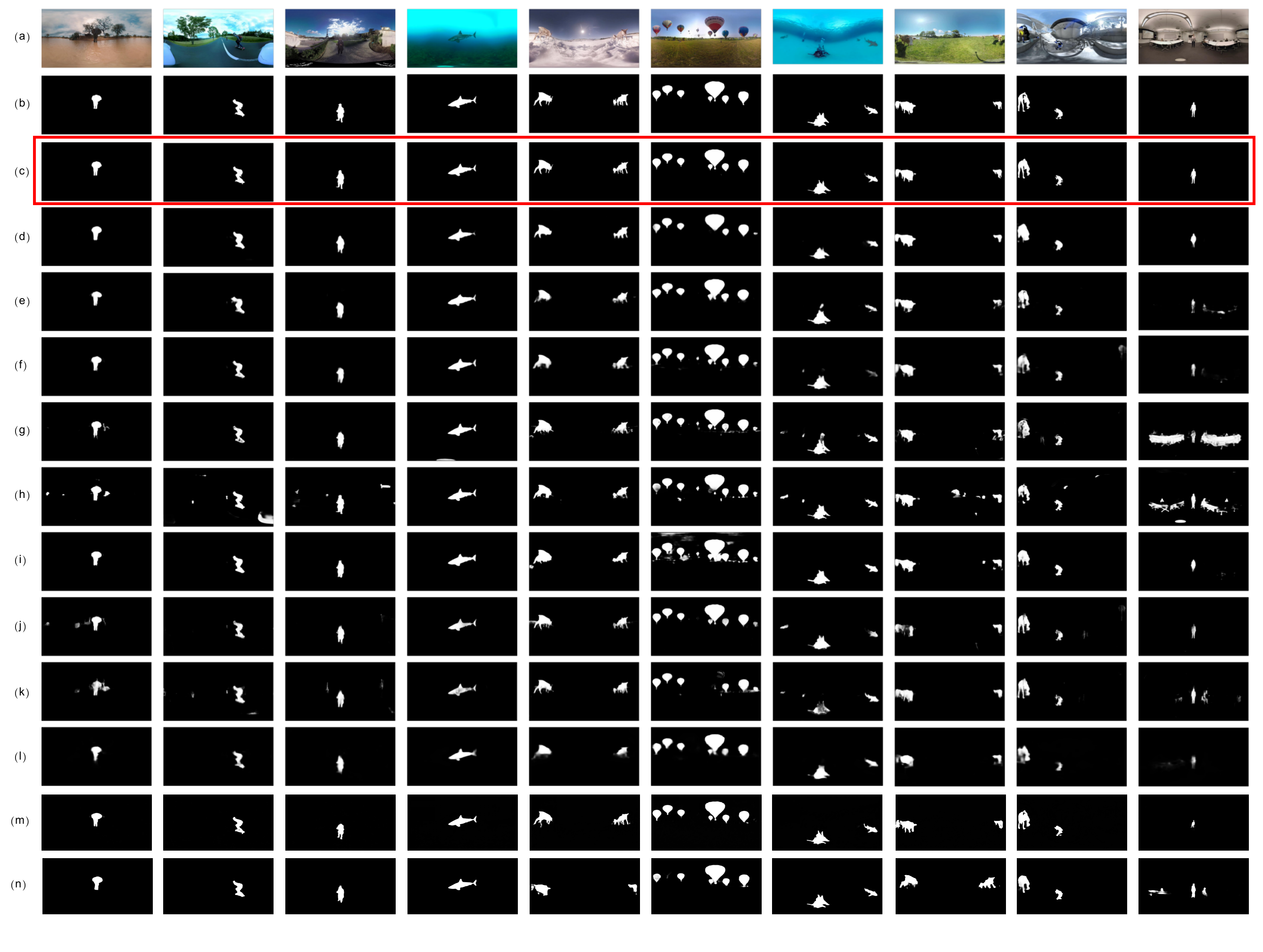}
	\caption{Visual comparisons of the different methods. (a) Equirectangular images. (b) Ground truth. (c) Result of our proposed method (marked in red box). (d)-(n) Saliency maps generated by FANet\cite{huang2020fanet}, DDS \cite{li2019distortion}, GCPANet \cite{chen2020global}, ITSD \cite{zhou2020interactive}, F3Net\cite{wei2020f3net}, MINet\cite{pang2020multi}, HVPNet \cite{liu2020lightweight}, SAMNet \cite{liu2021samnet}, CPDNet \cite{wu2019cascaded}, BPFINet\cite{chen2021bpfinet}, CTDNet\cite{zhao2021complementary}.}
	\label{fig:vis}
\end{figure*}

\section{Experiments}
%In this section, we first introduce the experimental settings, including the datasets, evaluation metrics, and implementation details. Then, the qualitative and quantitative comparisons with other state-of-the-art methods are presented in Section IV-B. At last, we evaluate the effectiveness of our model through the ablation studies in Section IV-C.
\subsection{Experimental settings}

%\subsubsection{Datasets}
We evaluate our proposed model on two SOD datasets for 360$^{\circ }$ omnidirectional images. 360-SOD dataset \cite{li2019distortion} contains $500$ images, which are all intercepted from the existing indoor and outdoor omnidirectional videos. For this dataset, $400$ images are used for training and the remaining $100$ images are used for testing.  F-360iSOD dataset \cite{zhang2020fixation} is a fixation-based dataset, which contains $107$ images with challenging real world daily scenes. All images in this dataset are used for testing.
%\subsubsection{Evaluation metrics}
To evaluate the performance of our proposed model, we calculate five evaluation metrics, including F-measure (${F}_{\beta }$) \cite{crm2019tcsvt}, weighted F-measure ($F_{\beta}^{\omega}$) \cite{margolin2014evaluate}, MAE \cite{crm2019tcsvt}, S-measure ($S_m$) \cite{fan2017structure}, and E-measure ($E_\Phi$) \cite{fan2018enhanced}.
We implement our network with Pytorch and train with an NVIDIA 3090 GPU (22G memory).  We also implement our network by using the MindSpore Lite tool\footnote{\url{https://www.mindspore.cn/}}. We adopt ResNet-50 \cite{he2016deep} pretrained on ImageNet \cite{deng2009imagenet} as backbone of our network and  set the dilation rates of the last two residual blocks to two for larger receptive as DeepLab\cite{chen2018encoder}. The equirectangular images are resized to $1024\times 512$, and the side length of the cube-map is $256$. Thus, the size of the vertical image and horizontal image of a CU image are $256\times 768$ and $1024\times 256$, respectively. During training, the batch size is set to $2$ and the learning rate is initiated to $5e^{-6}$. For the optimizer, we use ADAM with momentum of $0.9$, ${\beta }_{1} $, ${\beta }_{2} $, and $\varepsilon$ are set to $0.9$, $0.999$, and $10e^{-8}$ for network optimization, respectively.
%We stop the training process after 150 epochs (about 19.5 hours). The average inference time for a 360$^{\circ}$ omnidirectional image with a resolution of $1024\times 512$ image is 0.85s.

\subsection{Comparison with the State-of-the-arts}
In experiments, we compare our method with other fifteen state-of-the-art methods on the 360-SOD dataset and F-360iSOD datasets, including thirteen 2D SOD methods (\ie, SAMNet \cite{liu2021samnet}, HVPNet \cite{liu2020lightweight}, ITSD \cite{zhou2020interactive}, F3Net \cite{wei2020f3net}, GCPANet \cite{chen2020global}, MINet \cite{pang2020multi}, CPDNet \cite{wu2019cascaded}, R3Net \cite{deng2018r3net}, GFINet \cite{zhu2021supplement}, PFSNet \cite{ma122021pyramidal}, Auto-MSFNet \cite{zhang2021auto}, CTDNet \cite{zhao2021complementary}, BPFINet \cite{chen2021bpfinet} ) and two 360 SOD methods (\ie, DDS \cite{li2019distortion} and FANet \cite{huang2020fanet}). For a fair comparison, we retrained all the 2D SOD models on the 360-SOD dataset under the default parameters.

\subsubsection{Qualitative Comparison}
The visual results of different methods are showed in Fig. \ref{fig:vis}. As visible, our method achieves the best visual effects, with strong background suppression capabilities, complete foreground detection capabilities and elaborate detail representation capabilities. One one hand, compared with the retrained 2D SOD methods, the 360 SOD methods have more advantages. For example, in the fifth image, almost all methods cannot completely detect all the salient objects, either containing redundant background interference (\eg, GCPANet \cite{chen2020global}, MINet\cite{pang2020multi}) or missing some salient objects (\eg, the biggest hot air balloon in SAMNet \cite{liu2021samnet}). On the other hand, compared with other 360 SOD methods, the performance of our model is more competitive in both simple and complex scenarios.
For the simple scenes, our method is more accurate in details than other methods (\eg, the front legs of an elephant in the first image, the wheels of a skateboard in the second image, and the tail of fish in the sixth image). This shows the adaptability of our model to the distorted objects in 360$^{\circ}$ omnidirectional images.
For some more complex scenes, such as the the fifth image and ninth image, our method still achieves satisfactory results in terms of object completeness and detail description. In the fifth image, the FANet \cite{huang2020fanet} method wrongly suppressed the fourth small hot air balloon from the left, and neither the DDS \cite{li2019distortion} nor FANet \cite{huang2020fanet} methods correctly detected the nacelle. By contrast, these regions are all detected completely and accurately by our proposed method. Similarly, in the ninth image, the DDS \cite{li2019distortion} method fails to suppress the interference from the large white table area on the right, while the FANet \cite{huang2020fanet} method effectively suppresses the interference, but the boundary of the detected salient object is very blurred. Looking back on our method, it can clearly detect the arms, legs of the foreground regions while effectively suppressing the background regions.
\begin{table*}[!t]
\footnotesize
\renewcommand\arraystretch{1.1}
  \caption{Quantitative comparison on two public datasets. From top to bottom: SOD methods on 2D images, SOD methods on 360$^{\circ}$ omnidirectional images and our method. $\uparrow$ and $\downarrow$ denote larger and smaller is better. The top three results are marked in \textcolor[rgb]{1.000, 0.000, 0.000}{red}, \textcolor[rgb]{0.000, 1.000, 0.000}{green} and \textcolor[rgb]{0.000, 0.000, 1.000}{blue}.}
    \begin{center}
\begin{tabular}{c|c|c|ccccc|ccccc}

\toprule
\multirow{2}{*}{ }&\multirow{2}{*}{Year} &\multirow{2}{*}{Sources} &\multicolumn{5}{c}{360-SOD Dataset} &\multicolumn{5}{|c}{F-360iSOD Dataset}\\
\cline{4-13}
 & & & $E_\Phi \uparrow$ & $S_m \uparrow$ & $F_{\beta}^{\omega}\uparrow$ &  $F_{\beta}\uparrow$  &  MAE $\downarrow$ &  $E_\Phi \uparrow$ & $S_m \uparrow$ & $F_{\beta}^{\omega}\uparrow$ &  $F_{\beta}\uparrow$  &  MAE $\downarrow$ \\
\hline
    R3Net\cite{deng2018r3net}  & 2018 & IJCAI
            & 0.661 & 0.750 & 0.454 & 0.429 & 0.039
            & 0.704 & 0.512 & 0.247 & 0.284 & 0.069 \\
    CPDNet\cite{wu2019cascaded}  & 2019 & CVPR
            & 0.806 & 0.771 & 0.600 & 0.584 & 0.027
            & 0.746 & 0.580 & 0.324 & 0.358 & 0.063 \\
    MINet\cite{pang2020multi}  & 2020 & CVPR
           & 0.814 & 0.768 & 0.620 & 0.622 & 0.035
           & 0.676 & 0.563 & 0.307 & 0.336 & 0.068 \\
    GCPANet\cite{chen2020global}  & 2020 & AAAI
             & 0.841 & 0.805 & 0.662 & 0.632 & \textcolor[rgb]{0.000, 0.000, 1.000}{0.023}
             &  0.743     &  0.581    &  0.339     & \textcolor[rgb]{0.000, 0.000, 1.000}{0.375}  & 0.065 \\
    F3Net\cite{wei2020f3net}  & 2020 & AAAI
           & 0.778 & 0.730  & 0.515 & 0.532 & 0.037
           &  0.735     & 0.593 &   0.348   &  0.365     & 0.067\\
    ITSD\cite{zhou2020interactive}  & 2020 & CVPR
          & 0.808 & \textcolor[rgb]{0.000, 0.000, 1.000}{0.812} & 0.636 & 0.583 & 0.025
             & 0.726 & 0.597 & 0.367 & 0.370 &  0.075 \\
    HVPNet\cite{liu2020lightweight}  & 2020 & TCYB
            & 0.788 & 0.739 & 0.555 & 0.552 & 0.030
            &  0.716  &  0.534 &   0.222  &  0.311     &0.061 \\
    SAMNet\cite{liu2021samnet}  & 2020 & TIP
            & 0.716 & 0.759 & 0.516 & 0.473 & 0.037
            &  0.537&  0.525 & 0.224 & 0.202& 0.145 \\
    GFINet\cite{zhu2021supplement} & 2021	& TNNLS
           & 0.648	& 0.711	& 0.426	& 0.520	& 0.053 
           & 0.621	& \textcolor[rgb]{0.000, 1.000, 0.000}{0.615}	& 0.320	& 0.343	& 0.120\\
    PFSNet\cite{ma122021pyramidal} & 2021	& AAAI
           & \textcolor[rgb]{0.000, 1.000, 0.000}{0.885}	& 0.811	& \textcolor[rgb]{0.000, 0.000, 1.000}{0.711}	& \textcolor[rgb]{0.000, 1.000, 0.000}{0.731}	& \textcolor[rgb]{0.000, 1.000, 0.000}{0.021}
           & 0.717	& 0.606	& \textcolor[rgb]{0.000, 1.000, 0.000}{0.378}	& 0.360	& \textcolor[rgb]{0.000, 1.000, 0.000}{0.057}\\
   Auto-MSFNet\cite{zhang2021auto} & 2021	& ACM MM
                & 0.879	& 0.807	& 0.685	& 0.698	& \textcolor[rgb]{0.000, 0.000, 1.000}{0.023}
                & 0.739	& 0.578	& 0.371	& 0.364	& 0.062\\
    CTDNet\cite{zhao2021complementary} & 2021	& ACM MM
           & \textcolor[rgb]{0.000, 0.000, 1.000}{0.883} & 0.806 & 0.694 & \textcolor[rgb]{0.000, 0.000, 1.000}{0.714}	& \textcolor[rgb]{0.000, 0.000, 1.000}{0.023}
           & \textcolor[rgb]{0.000, 0.000, 1.000}{0.749}	& \textcolor[rgb]{0.000, 0.000, 1.000}{0.614}	& \textcolor[rgb]{0.000, 0.000, 1.000}{0.377}	& 0.367	& \textcolor[rgb]{0.000, 0.000, 1.000}{0.059}\\
    BPFINet\cite{chen2021bpfinet} & 2021	& NC
            & 0.875	& 0.794	& 0.672	& 0.700	 & 0.027
            & 0.743	& 0.608	& 0.374	& 0.356	& 0.065\\
    
\hline
    DDS\cite{li2019distortion}    & 2020 & STSP
           & 0.854 & 0.799 & 0.663 & 0.638 & \textcolor[rgb]{0.000, 0.000, 1.000}{0.023}
            & \textcolor[rgb]{0.000, 1.000, 0.000}{0.760}  & 0.577 & 0.295 &0.364 & 0.061 \\
    FANet\cite{huang2020fanet}  & 2020 & SPL
           & \textcolor[rgb]{0.000, 0.000, 1.000}{0.883} & \textcolor[rgb]{0.000, 1.000, 0.000}{0.826} & \textcolor[rgb]{0.000, 1.000, 0.000}{0.717} & 0.700   & \textcolor[rgb]{0.000, 1.000, 0.000}{0.021}
        & 0.747 & 0.587 & 0.355 & \textcolor[rgb]{0.000, 1.000, 0.000}{0.381} & 0.061 \\
    MPFR-Net (Ours)  &-&-& \textcolor[rgb]{1.000, 0.000, 0.000}{0.892} & \textcolor[rgb]{1.000, 0.000, 0.000}{0.842} & \textcolor[rgb]{1.000, 0.000, 0.000}{0.744} & \textcolor[rgb]{1.000, 0.000, 0.000}{0.754} & \textcolor[rgb]{1.000, 0.000, 0.000}{0.019} & \textcolor[rgb]{1.000, 0.000, 0.000}{0.762} & \textcolor[rgb]{1.000, 0.000, 0.000}{0.617} & \textcolor[rgb]{1.000, 0.000, 0.000}{0.392} & \textcolor[rgb]{1.000, 0.000, 0.000}{0.430} & \textcolor[rgb]{1.000, 0.000, 0.000}{0.052} \\
\bottomrule
\end{tabular}
  \label{tab:pf1}
  \end{center}
\end{table*}
\subsubsection{Quantitative Comparison}
As for quantitative comparison, we use five metrics to evaluate different methods on two datasets, which is reported in Table \ref{tab:pf1}. On the 360-SOD dataset \cite{li2019distortion}, our model achieves the best performance in all five evaluation metrics against other comparison models. For example, the F-measure is increased from $0.731$ to $0.754$ compared with the \textbf{second best} result, with the percentage gain of $3.15$\%. Moreover, the MAE score wins the percentage gain of $9.52$\% compared with the \textbf{second best} result. On the F-360iSOD dataset, compared to other algorithms, we achieve the best performance in all five evaluation metrics expect the weighted F-measure. Compared with the \textbf{second best} result, the MAE score is updated form $0.057$ to $0.052$ with the percentage gain of $8.77$\%. In general, our method achieves very competitive performance in both quantitative and qualitative evaluation, which also demonstrates the effectiveness of our network design. Besides, we show the comparisons of computation time  in Table \ref{tab:time}. Our method takes $0.39$ second to process an image with the size of $1024 \times 512$, which roughly ranks in the middle of the comparison algorithms.

\begin{table}[!t]
\footnotesize
\renewcommand\arraystretch{1.1}
	\caption{Average computation time of different methods on the 360-SOD dataset.}
	\centering
		\begin{tabular}{c|c|c|c|c} 
		\toprule
		Method	&R3Net	&CPDNet	&MINet	&GCPANet\\
		\hline
		Time (s)	&0.42	&0.28	&0.41	&0.44\\
		\hline
		Method	&F3Net	&ITSD	&HVPNet	&SAMNet\\
		\hline
		Time (s)	&0.43	&0.40	&0.34	&0.27\\
		\hline
		Method	&PFSNet	&Auto-MSFNet	&CTDNet	&BPFINet\\
		\hline
		Time (s)	&0.28	&0.47	&0.71	&0.35\\
		\hline
		Method	&GFINet	 &DDS	&FANet	&Ours\\
		\hline
		Time (s)	&0.38	&-	&0.21	&0.39\\
		\bottomrule
        \end{tabular}%
	\label{tab:time}	
\end{table}		
\subsection{Ablation Study}
In this section, we conduct comprehensive ablation study to verify the design of the proposed method by disabling the corresponding components respectively. The quantitative results on the 360-SOD dataset are reported in Table \ref{tab:ablation}. The visual results are shown in Fig. \ref{fig:abl}. In the table, `$w/o$ CU' refers to only use equirectangular image as input without cube-unfolding images, `$w/o$ DWF' refers to replace the DWF module by directly adding the multi-projection features together, `$w/o$ FR' refers to replace the FR module by concatenation operation, `$w/o$ WAF' refers to replace the WAF module by allocatting equal weights, and `$w/$ six cube surfaces' refers to replace the input of the network from four cube-unfolding images to the six cube surfaces.

\subsubsection{Cube unfolding}
\begin{figure*}[!t]
 	\centering
 	\includegraphics[width=\textwidth]{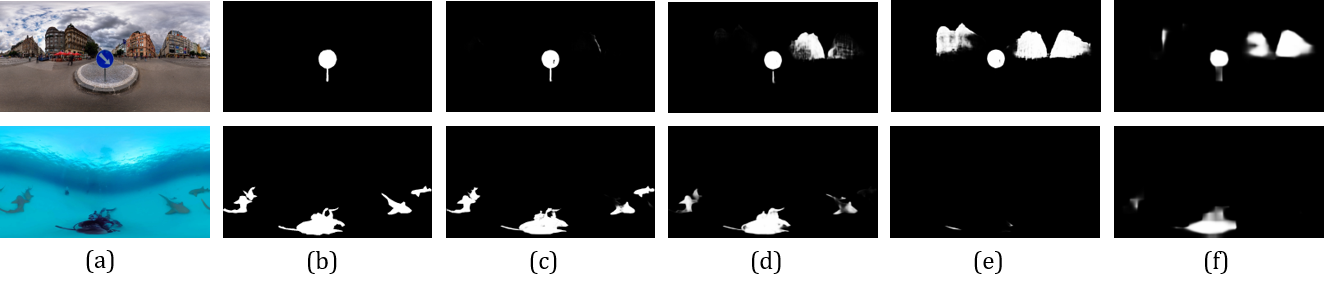}
 	\caption {Visual results of ablation study. (a) Equirectangular images, (b) Ground truth, (c)-(f) Results of full model, $w/o$ CU, $w/o$ DWF and $w/o$ FR.}
 	\label{fig:abl}
 \end{figure*}

In the proposed model, we combine multiple projection manners of the 360$^{\circ}$ omnidirectional images, including an equirectangular projection (EP) image and four cube-unfolding (CU) images.
In order to prove the effectiveness of introducing CU images, we remove the CU images and only use the EP image as input. The result is denoted as `$w/o$ CU' in Table \ref{tab:ablation}.
Compared with the full model, after removing the CU inputs, the performance of all indicators is reduced. For example, the F-measure decreased by $4.14$\% from $0.754$ to $0.724$. As shown in Fig. \ref{fig:abl}, when only EP image is used, the detection result cannot well suppress the interference in the background regions (such as the building in the first image), and it will also lead to missed detection (such as the shark on the right in the second image).
To further prove the benefit of introducing CU images, we replace the input of the network from four cube-unfolding images to the six cube surfaces as in \cite{huang2020fanet}, maximally preserving the original modules for fair comparison. Specifically, in DWF module, the features of six cube surfaces are transformed into one equirectangular feature through a cube-map to equirectangular projection, and then fused with the original EP branch feature through the GEF unit. Furthermore, as the role of the WAF unit is to assign weights to the four cube unfolding features in our network, and only one cube can be formed from the six cube surfaces in this ablation experiment, so the WAF unit is not needed. The rest of the network remains the same. The result is denoted as `$w/$ six cube surfaces' in Table \ref{tab:ablation}. From it, we can see that our model with the cube-unfolding achieves better performance than the our model with six cube surfaces. For example, the F-measure is increased from 0.732 to 0.754 with the percentage gain of 3.0\%, and the MAE score is improved from 0.021 to 0.019 with the percentage gain of 9.5\%. All these results show that the effectiveness of the cube unfolding representation method.

Besides, we also discuss the network performance when the CU images are input in different orders. In our network, four CU images are embedded into the feature extraction backbone in a parallel manner. Furthermore, in subsequent modules, all processing operations (\eg, DWF, FR) on the four cube-unfolding features are parallelized and adaptive. Therefore, in theory, all operations in our network are order-invariant. To verify this, we randomly select three orders from 24 possibilities for verification. Assuming that the original order of the cube-unfolding images is represented as 1-2-3-4, then the three sampled orders in this experiment can be represented as 1-3-2-4, 2-4-3-1, and 4-2-1-3. The rest of the network and the training parameter settings remain unchanged, and the results are shown in Table \ref{tab:inputorder}. As can be seen, the performance of the three randomly sampled orders is very close to the performance of the original order, indicating that our model is insensitive to the order of four CU images. 

\begin{table}[!t]
\footnotesize
\renewcommand\arraystretch{1.1}
	\caption{Ablation studies on the 360-SOD dataset. }
	\centering
		\begin{tabular}{c|ccccc}
        \toprule
  %      \multicolumn{6}{c}{360-SOD} \\
  %      \midrule
              &  $E_\Phi \uparrow$ & $S_m \uparrow$ & $F_{\beta}^{\omega}\uparrow$ &  $F_{\beta}\uparrow$  &  MAE $\downarrow$ \\
        \midrule
        Full model  & \textbf{0.892} & \textbf{0.842} & \textbf{0.744} & \textbf{0.754} & \textbf{0.019}\\
        \midrule
        $w/o$ CU & 0.877 &  0.834  &  0.731 & 0.724 &  0.021\\
        \midrule
        $w/o$ DWF & 0.873 & 0.828 & 0.728 & 0.740 & 0.021 \\
        \midrule
        $w/o$ FR & 0.829 & 0.809 & 0.645 & 0.617 & 0.024 \\
        \midrule
        $w/o$ WAF & 0.875 &0.833 & 0.731 &0.743	& 0.021 \\
        \midrule
        $w/$ six cube surfaces	&0.878	&0.831	&0.707 &0.732	&0.021\\

        \bottomrule
        \end{tabular}%
	\label{tab:ablation}	
\end{table}

\subsubsection{Dynamic weighted fusion}
The DWF module is used to integrate the multiple projection features through learnable dynamic weights from the perspective of intra fusion and inter fusion. In order to show the effectiveness of the DWF module, we replace the dynamic weighted fusion strategy with a normal additional fusion operation, which is denoted as `$w/o$ DWF' in Table \ref{tab:ablation}. To be fair, the two methods set the same number of convolutional layers.
We can see that after removing the DWF module, the performance is significantly reduced. For example, both the S-measure and the F-measure dropped by 1.4\%.
The visualization result shown in Fig. \ref{fig:abl} once again prove the effectiveness of DWF. Even if EP image and CU images are used as input at the same time, if there is no reasonable information fusion method, it may lead to poor detection results. For example, the building backgrounds in the first image are wrongly reserved, and the second image does not even detect any effective salient targets at all.
In addition, as the WAF unit is an important unit in the DWF module, we also add an ablation experiment of the WAF unit in Table \ref{tab:ablation} (denoted as `$w/o$ WAF'), replacing it by allocating equal weights for the four CU features. From it, we can see that after removing the WAF unit, the performance on most evaluation metrics is reduced. For example, the E-measure is decreased from 0.892 to 0.875, with a reduction of 1.7\%.

\begin{table}[!t]
\footnotesize
\renewcommand\arraystretch{1.1}
	\caption{Quantitative results with different CU images input orders.}
	\centering
		\begin{tabular}{c|ccccc}
        \toprule
              &  $E_\Phi \uparrow$ & $S_m \uparrow$ & $F_{\beta}^{\omega}\uparrow$ &  $F_{\beta}\uparrow$  &  MAE $\downarrow$ \\
        \midrule
        1-2-3-4 (original order)  & 0.892 & 0.842 & 0.744 & 0.754 & 0.019\\
        \midrule
        1-3-2-4 & 0.889 &  0.841  &  0.743 & 0.751 &  0.020\\
        \midrule
        2-4-3-1 & 0.888 & 0.844 & 0.746 & 0.757 & 0.018 \\
        \midrule
        4-2-3-1 & 0.893 & 0.843 & 0.743 & 0.754 & 0.020 \\
        \bottomrule
        \end{tabular}%
	\label{tab:inputorder} 
\end{table}

\subsubsection{Filtration and refinement}
The FR module is designed for the combination of features from different levels, by filtrating the redundant information of features and refining the high-level semantic feature. To demonstrate the effectiveness of the FR module, we replace it with a simple concatenation operation, and the result is shown as `$w/o$ FR' in Table \ref{tab:ablation}. Similarly, for fairness, we let both methods go through the same number of convolution layers.
As can be seen, the filtration and refinement strategies proposed in this paper achieves obvious performance advantages in all evaluation indicators. For example, after introducing the FR module, the F-measure reaches a performance gain of 22.20\%, and the MAE score win a gain of 20.83\%. Compared with the full model in Fig. \ref{fig:abl}, after removing the FR module, the background suppression capability (\eg, the buildings in the first image) the completeness and accuracy of the detected salient objects (\eg, the missing sharks in the second image) are significantly degenerated.

Although the FR module is also used to combine the encoder and decoder features, just like the FIA module in \cite{chen2020global}, but they have distinct differences. 
On the one hand, due to the multi-branch structure of the encoder, FR module combines the decoder features and five encoder features from the EP branch and CU branch. Instead, the FIA module in \cite{chen2020global} integrates the decoder features and single encoder features with the additional global contextual features. On the other hand, the biggest difference from FIA module in \cite{chen2020global} is that we use a serial stepping structure, while FIA module in \cite{chen2020global} is a mirror-symmetric parallel structure. Concretely, FIA module in \cite{chen2020global} uses the original low-level features to supplement the high-level features in detail, while using the original high-level features to guide the low-level features by providing semantic information, which is a complete mirroring strategy. However, our FR module adopts a step-by-step strategy, first enhancing decoder features, then filtering the five encoder features with the enhanced decoder features, and further refining decoder features with the filtrated encoder features. In this way, we can suppress the redundant background information while retaining details relevant to the salient object. In order to verify the effect of our FR module, we replace the combination of low-level features and high-level features in the FIA module of \cite{chen2020global} with the step-by-step strategy in our FR module. The results on the ECSSD \cite{shi2015hierarchical} dataset are shown in Table \ref{tab:fr}. According to the quantitative results, it can be seen that, except for the MAE score, other indicators have been improved to a certain extent after the replacement. This verifies the effectiveness of design of progressive fusion in the FR module.
\begin{table}[t]
\footnotesize
\renewcommand\arraystretch{1.1}
	\caption{Quantitative evaluation of FR module in GCPANet \cite{chen2020global} on the ECSSD \cite{shi2015hierarchical} dataset.}
	\centering
		\begin{tabular}{c|ccccc}
        \toprule
  %      \multicolumn{6}{c}{360-SOD} \\
  %      \midrule
              &  $E_\Phi \uparrow$ & $S_m \uparrow$ & $F_{\beta}^{\omega}\uparrow$ &  $F_{\beta}\uparrow$  &  MAE $\downarrow$ \\
        \midrule
        GCPANet $w/$ FR	&0.923	&0.927	&0.904	&0.919 &0.036\\
        \midrule
        GCPANet	&0.921	&0.925	&0.899	&0.916	&0.036\\
        \bottomrule
        \end{tabular}%
	\label{tab:fr}
\end{table}
\subsection{Discussion}
%\textcolor{red}{Although the proposed approach achieves a competitive performance, there are still some limitations. Future work can be further explored in terms of the following aspects.}

\begin{figure}[!t]
 	\centering
 	\includegraphics[width=\columnwidth]{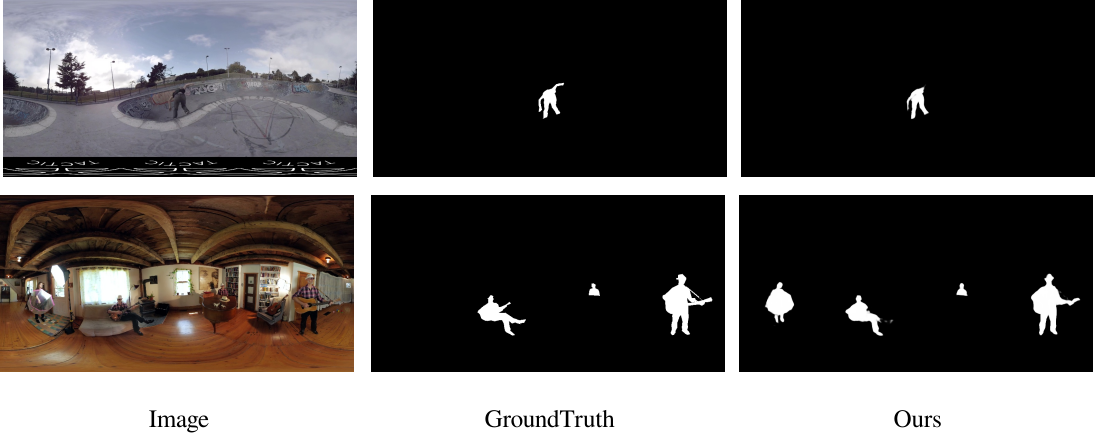}
 	\caption {Failure cases of our method.}
 	\label{fig9}
 \end{figure}

\subsubsection{Failure cases}
Due to the large and complex background scene of the 360$^{\circ}$ omnidirectional image, objects occupy a relatively small proportion in any cases. Especially when the contrast between the object and the background is low, the task is even more challenging. For example, in the first image of Fig. \ref{fig9}, the skateboarder has an outstretched hand hidden in the background. Our method fails to detect the pose of his hands in their entirety. Besides, in the second image, the legs of the seated man and the guitar strap on the far are easily suppressed as background regions. Therefore, it is a challenge to strike a balance between preserving the independent details of the object and suppressing background redundancy. In addition, as shown in the second image, the little girl holding an umbrella on the left is wrongly detected as the salient object by our method. It is also a challenge to determine whether an object is salient or not in the context of a cluttered scene with many objects. For such problems, we can resort to stronger feature extractors (such as swin Transformer \cite{liu2021swin}) and some reasoning mechanisms (such as GNN \cite{scarselli2008graph}) in the future.

\subsubsection{Future work}
To facilitate the storage and processing of omnidirectional images, the equirectangular projection method has been widely used to convert an omnidirectional image into a 2D image, which is closer to the human perceptual mechanism. Moreover, the final performance evaluation is also carried out on the equirectangular projection image, which has become a mainstream practice. However, any projection method for omnidirectional images inevitably introduces distortion, including our proposed cube-unfolding representation, but it can bring less distortion than the equirectangular projection while preserving object integrity as much as possible. In other words, dealing with the artifacts caused by the projection distortion has become one of the challenges for omnidirectional image related tasks. In contrast, the omnidirectional image is free from distortion in the spherical domain. However, implementing SOD in the spherical domain is much more difficult because it is completely different from the traditional 2D image processing and requires extending the SOD from 2D space to 3D space. To address this challenge, some works have redesigned the convolution kernels specifically for spheres, extending the CNN from traditional 2D grid to spherical domain \cite{coors2018spherenet}. In addition, there are other projection domains for omnidirectional images that are closer to spheres, such as dodecahedron \cite{chen2018recent} and icosahedron \cite{lee2019spherephd}, but these projection domains are relatively rare in practical applications. In the future, fully exploring the impact of different omnidirectional image projection methods on SOD task and choosing an effective and efficient representation manner are issues worthy of in-depth consideration, such as the combination of spherical and equirectangular projections.

In addition, as shown in Table \ref{tab:time}, due to the high resolution and wide scene range of omnidirectional images, the existing SOD methods usually take more than 0.2 seconds to processing an omnidirectional image, including some SOD models specifically designed for real-time applications. Although our proposed model has advantages in accuracy and detection effect, its inference time is still unsatisfactory and insufficient to support some real-time scene applications. In the future, on the one hand, we can improve the model by using more lightweight backbone and streamlining the modules. On the other hand, we need to further explore efficient feature representation methods for omnidirectional images, and can borrow some ideas from SOD methods for high-resolution images \cite{zeng2019towards,tang2021disentangled}.

\section{Conclusion}
In this paper, we propose an end-to-end salient object detection network for 360$^{\circ }$ omnidirectional images, named Multi-Projection Fusion and Refinement (MPFR) network. We consider combining the equirectangular projection and the cube-map projection to take full advantages of these two projection manners from sphere to plane. We propose a strategy to express the cube-map projection using four different cube-unfolding images, and take both equirectangular projection (EP) image and the four cube-unfolding (CU) images as inputs of the network. In the proposed MPFR network, a Dynamic Weighted Fusion (DWF) module is designed to integrate the features of multi-projection in a dynamic manner, including the inter fusion between the EP and each CU features, and the intra fusion between the four different CU-induced features. In addition, a Filtration and Refinement (FR) module is proposed to filtrate the redundant information of multi-level and multi-projection features, and refine the high-level semantic features. Experimental results on two 360$^{\circ }$ image SOD datasets demonstrate that the proposed model achieves the competitive performance compared with other state-of-the-art methods.
%-------------------------------------------------------------------------

\par
\ifCLASSOPTIONcaptionsoff
  \newpage
\fi
{
\bibliographystyle{IEEEtran}
\bibliography{egbib}
}
\end{document}